\documentclass{article} 
\usepackage{times}

\usepackage[preprint]{neurips_2022}

\usepackage{amsmath,amsfonts,bm}

\def\eqref#1{equation~\ref{#1}}

\def\1{\bm{1}}

\DeclareMathAlphabet{\mathsfit}{\encodingdefault}{\sfdefault}{m}{sl}
\SetMathAlphabet{\mathsfit}{bold}{\encodingdefault}{\sfdefault}{bx}{n}

\DeclareMathOperator*{\argmin}{arg\,min}

\usepackage{hyperref}
\usepackage{url}

\usepackage{booktabs}       
\usepackage{amsfonts}       
\usepackage{nicefrac}       
\usepackage{microtype}      
\usepackage{xcolor}         

\usepackage{amsmath}
\usepackage{graphicx}
\usepackage{subfigure}
\usepackage{multirow}
\usepackage{bbding}
\usepackage[noend]{algpseudocode}
\usepackage{algorithm}
\usepackage{algorithmicx}

\usepackage{enumitem}

\newtheorem{theorem}{Theorem}
\newtheorem{assumption}{Assumption}

\usepackage{colortbl}
\usepackage{diagbox}
\definecolor{mygray}{gray}{.9}

\title{Exploiting Personalized Invariance for Better Out-of-distribution Generalization in Federated Learning}

\author{Xueyang Tang, Song Guo, Jie Zhang\\
Department of Computing\\
The Hong Kong Polytechnic University\\
\texttt{\{xyang.tang,jieaa.zhang\}@connect.polyu.hk,song.guo@polyu.edu.hk}\\
}

\begin{document}

\maketitle

\begin{abstract}

  Recently, data heterogeneity among the training datasets on the local clients (a.k.a., Non-IID data) has attracted intense interest in Federated Learning (FL), and many personalized federated learning methods have been proposed to handle it. However, the distribution shift between the training dataset and testing dataset on each client is never considered in FL, despite it being general in real-world scenarios. We notice that the distribution shift (a.k.a., out-of-distribution generalization) problem under Non-IID federated setting becomes rather challenging due to the entanglement between personalized and spurious information.
  To tackle the above problem, we elaborate a general dual-regularized learning framework to explore the \textit{personalized invariance}, compared with the exsiting personalized federated learning methods which are regularized by a single baseline (usually the global model).
  Utilizing the personalized invariant features, the developed personalized models can efficiently exploit the most relevant information and meanwhile eliminate spurious information so as to enhance the out-of-distribution generalization performance for each client. Both the theoretical analysis on convergence and OOD generalization performance and the results of extensive experiments demonstrate the superiority of our method over the existing federated learning and invariant learning methods, in diverse out-of-distribution and Non-IID data cases.
\end{abstract}

\section{Introduction}
\label{intro}

As data privacy is attached more and more importance to in the field of machine learning, federated learning~\citep{mcmahan17} (FL) gains increasing attention and dramatic development in recent years.
Federated learning allows the participation of a massive number of data holders (usually called clients) to train the learning models in a collaborative manner. Most importantly, the participating clients can preserve their data locally during the collaborative training process, and hence prevent the leakage of private data. Traditional federated learning approaches develop a shared gloal model by model aggregation to fit all the local datasets, which can work well when the local data instances on clients subjects to independent and identically distribution (IID). Conversely, the Non-IID setting where the local datasets don't satisfy the IID assumption can degrade the performance (both model accuracy and convergence rate) of the global model dramatically~\citep{hsieh20noniid}. To settle the challenge of Non-IID data, personalized federated learning (PFL) is proposed to train a personalized model for each client. Contrary to the individually trained local model which is often trapped in overfitting due to the insufficient local data, the personalized model obtains some generalization knowledge from the sever or the cooperation with other clients. 

Despite the exsiting personalized federated learning methods achieve great success recently, we notice that all of them focus on ,the model performance on in-distribution testing dataset. In this way, the produced models can hardly adapt to the scenarios where the testing dataset is out-of-distribution in terms of the training dataset. Here, we refer to out-of-distribution (OOD) scenario as the case where there exists distribution shift between the training and testing dataset. In fact, the out-of-distribution scenario is more genreal and practical in machine learning applications, and has drawn lots of efforts in the literature on OOD generalization~\citep{arjovsky2019IRM,sagawa2019DRO,creager2021EIIL,ye2021theoretical,shen2021surveyOOD,krueger2021out,zhang2021deep}. However, the out-of-distribution generalization problem is mostly studied in centralized setting where all data can be accessed, but has seldom been considered in the federated learning. Although some of the methods designed for centralized setting can be naively applied into the federated learning to generate a shared global model, they cannot cast off the Non-IID data quagmire because the gained model drops the useful information about personalization. In summary, the existing federated learning works can hardly handle the out-of-distribution scenarios, and meanwhile the centralized methods for OOD generalization will be troubled by the Non-IID data even if they can be implemented in federated learning setting.

What's worse, we find that the simple combination of the existing PFL and OOD methods, instead of solving the OOD generalization problem under Non-IID federated setting, will produce much worse performance than the better one of themselves (the finding will be shown and discussed in the experiment part). To handle the challenging OOD problem under Non-IID federated setting, we disscting the data heterogeneity (Non-IID) from two orthogonal perspectives and then introduce a novel concept: \textit{personalized invariance}. The proposed personalized invariance can preserve the personal information, and in the meantime eliminate the spurious information to equip the obtained personalized models with out-of-distribution generalization.
In contrast with the naive combination of PFL and OOD methods which can easily `pick up' the spurious correlation during the process of personalization because of the entanglement of personalized and spurious information, we propose a general learning framework that is regularized by two baselines: global invariance and local invariance. The global invariance can prevent the personalized models from being stuck in overfitting due to the insufficiency of local data. However, global invariance will discard both personalized and spurious information. Therefore, the constraint of local invariance is utilized to exclude the spurious correlation. With the guidance of the designed dual-constraint, the personalized models can effectively exploit personalized invariance to solve the challenging OOD problem for every client in federated setting. 
The main contributions of this paper are summarized as follows:
\begin{itemize}
	\item To the best of our knowledge, we are the first to consider and handle the challenging personalized out-of-distribution problem under Non-IID federated learning setting. We propose the novel concept of \textit{personalized invariance} and theoretically analyze the advantage of personalized invariance over the naive global invariance.
	\item We elaborate a dual-regularized learing framework to explore the personalized invariant features which can include important personalization information and meanwhile exclude the spurious correlation. Formally, the OOD problem under Non-IID setting is formulated as a bi-level optimization where the client-level objective is constrained by a dual-baseline regularization. Besides, we provide theoretical analysis on convergence rate and OOD generalization performance which proves the effectiveness of our method.
	\item We conduct extensive experiments on diverse datasets to compare the performance of our method with the existing federated learning (including personalized federated learning) and invariant learning methods. The results show that our method outperforms the state-of-the-art ones in varied out-of-distribution and Non-IID settings.
\end{itemize}

\section{Related work}
\label{related}

\subsection{Personalized Federated Learning}
One of the most challenging problems in federated learning~\citep{mcmahan17} is the data heterogeneity (a.k.a, Non-IID) across the local clients~\citep{hsieh20noniid}. To handle the Non-IID data quagmire, personalized federated learning (PFL) methods have been widely explored in many real-world applications.
Jeong et al. \citep{jeong2018communication,duan2020self} focus on data augmentation methods by generating additional data to augment its local data towards yielding an IID dataset. However, these methods usually require the FL server to know some statistical information about the local data distributions (e.g., class sizes, mean and standard deviation), which may potentially violate privacy policy \citep{tan2021towards}. Another line of work designs special client selection mechanisms to simulate homogeneous data distribution \citep{wang2020optimizing,yang2020federated,lyu2020towards}.

On the other hand, many model-based PFL methods focus on producing customized model structures or parameters for different clients, which can also be divided into two types: single-model, multiple-model based approaches. 
Single-model based methods extended from the conventional FL algorithms like FedAvg \citep{mcmahan17} combine the optimization of the local models and global model, including: local fine-tuning \citep{wang2019federated,schneider2019personalization,arivazhagan2019federated}, regularization \citep{dinh20pFedMe,hanzely20mix,hanzely20lower}, model mixture
\citep{mansour2020three,deng20peradap},  meta learning \citep{DBLP:conf/nips/0001MO20,jiang2019improving} and parameter decomposition \citep{bui2019federated,DBLP:conf/icml/CollinsHMS21,arivazhagan2019federated}. %
Considering the diversity and inherent relationship of local data, multi-model-based approaches where multiple global models are trained for heterogeneous clients are more reasonable. 
Some researchers \citep{DBLP:conf/nips/GhoshCYR20,mansour2020three,tangxCGPFL2022} propose to train multiple global models at the server, where similar clients are clustered into a group to share the in-distribution generalization knowledge. 
Another strategy is to collaboratively train a personalized model without explicit global models, e.g., FedAMP \citep{huang2021personalized}, FedFomo \citep{zhang2020personalized}, MOCHA \citep{DBLP:conf/nips/SmithCST17}, and KT-pFL \citep{zhang2021parameterized}. However, all of these works focus on the discrepancy between training data distributions rather than that between training and testing distributions.

\subsection{Out-of-distribution Generalization}
Out-of-Distribution (OOD) generalization problem refers to a challenging setting where the testing distribution is unknown and different from the training. In order to deal with the OOD generalization problem, tremendous efforts have been made and vary greatly ranging from causality to representation learning and from
structure-based to optimization-based. The existing methods can be categorized into three parts: unsupervised representation learning (e.g., to analyse causal relationships between data)~\citep{scholkopf2021toward,shen2020disentangled,yang2021causalvae,trauble2021disentangled,locatello2019challenging}, supervised model learning (e.g., design various model architectures and learning strategies to achieve OOD generalization, including domain generalization~\citep{wang2021generalizing,zhao2020domain,garg2021learn,du2020learning}, causal \& invariant learning~\citep{gamella2020active,oberst2021regularizing,krueger2021out,kamath2021does,arjovsky2019IRM} and stable learning~\citep{shen2018causally,shen2020stable,zhang2021deep}) and optimization approaches (focus on distributed robust optimization across different distributions~\citep{liu2021stable,sagawa2019DRO} or capturing invariant features~\citep{liu2021HRM,chang2020invariant,koyama2020out}).   
All of the above OOD generalization methods are studied in centralized scenarios, where all training data can be accessed. In some real-world applications, the data are usually generated locally and the data owners are not willing to share data with others due to the concern about privacy. Therefore, we are motivated to investigate the OOD generalization problem under the federated setting.


\section{Problem formulation}
\label{formulation}
In this section, we first introduce the preliminary knowledge of invariant learning
%
and then deconstruct the data heterogeneity into two orthogonal levels to investigate what information is necessary and what is spurious for solving the OOD problem under the Non-IID federated learning. 
For the purpose of fully exploiting the useful invariant information, we propose the personalized invariance which retains the important personalized information and concurrently excludes the spurious information to equip the obtained models with both personalization and out-of-distribution generalization. 
%

\paragraph{Notations.} Let $\mathcal{X}$ denotes the input space, $\mathcal{Y}$ denotes the target space, and correspondingly $(X, y) \in (\mathcal{X}, \mathcal{Y})$ is the data instance. The sets of training and testing environments are represented by $\mathcal{E}_{train}$ and $\mathcal{E}_{test}$ respectively. We use $\mathcal{E}_{all}$ as the set of all possible environments in the task concerned, i.e.,  $\mathcal{E}_{train}$, $\mathcal{E}_{test} \subset \mathcal{E}_{all}$. Suppose that there are $N$ clients and the local dataset $D_i, i \in [N]$ contains $m_i$ data instances on client $i$. For convenience, we separate the learning model or parameterized mapping from the input space $\mathcal{X}$ to $\mathcal{Y}$ into two consecutive parts: the feature extractor $\Phi$ and the classifier $w$. Specifically, the feature extractor $\Phi$ maps from the input space $\mathcal{X}$ to the latent feature space $\mathcal{H}$, i.e., $\Phi(X)\in \mathcal{H}$, and the classifier $w$ generates a hard prediction $\hat{y}$ from a latent feature $\Phi(X)$. Therefore, the overall learning model is denoted by $f_{\theta}=\Phi \circ w$, where $f_{\theta}$ indicates the function $f$ parameterized by $\theta$. In this paper, we define the expected empirical loss on dataset $D$ as $\mathcal{R}(f_{\theta};D):=\mathbb{E}_{(X,y)\in D}[\ell(f_{\theta}(X), y)]$, where $\ell$ is the loss function.

\subsection{Invariant Learning}
Invraint learning has been emerged as a promising approach for handling OOD generalization problem~\citep{arjovsky2019IRM,ahuja2020IRMgames,liu2021HRM,liu2021KernelHRM,creager2021EIIL}. The prevalent invariant learning assumes that there exists some invariant feature $\Phi(X)$ satisfying the \textbf{invariance property}:
\begin{small}
\begin{equation}
\mathbb{E}[y|\Phi(X) = h, e] = \mathbb{E}[y|\Phi(X) = h, e^{\prime}], \forall h \in \mathcal{H}, \forall e, e^{\prime} \in \mathcal{E}_{all} \tag{Invariance}
\label{invariance}
\end{equation}
\end{small}To discover the invariant representation $\Phi: \mathcal{X} \rightarrow \mathcal{H}$ that elicits an invariant predictor $w\circ \Phi$ across all environments, \textit{IRM}~\citep{arjovsky2019IRM} is proposed as a constrained optimization problem based on the empirical data from multiple accessible training environments $\mathcal{E}_{train}$. Out of efficiency concerns, a practical version of \textit{IRM} is offered in that work:
\begin{small}
\begin{equation}
\min_{f_{\theta}:\mathcal{X}\rightarrow \mathcal{Y}}\sum_{e \in \mathcal{E}_{train}}\mathcal{R}(f_{\theta};e) + \lambda \cdot{\lVert\nabla_{\bar{w}|\bar{w}=1.0}\mathcal{R}(\bar{w}\cdot f_{\theta};e)\rVert}^2, \tag{IRMv1}
\label{IRMv1}
\end{equation}
\end{small}where $\mathcal{R}(f_{\theta};e)$ is the expected empirical loss on environment $e$. In particular, $f_{\theta}$ is used as the entire invariant predictor, and $\bar{w}$ is a scalar and fixed "dummy" classifier as in~\citep{arjovsky2019IRM}. 
We define that $\mathcal{L}_{IRM}(\theta; \mathcal{E}_{train})=\sum_{e \in \mathcal{E}_{train}}\mathcal{R}(f_{\theta};e) + \lambda \cdot{\lVert\nabla_{\bar{w}|\bar{w}=1.0}\mathcal{R}(\bar{w}\cdot f_{\theta};e)\rVert}^2$ in this paper.

Apart from \textit{IRM}, another branch of approaches to OOD problem is worst-case generalization~\citep{sagawa2019DRO}. The target of worst-case generalization is to minimize the expected risk of the training enviroment which produces the worst risk such that the generated predictor can be simultaneously optimal for all environments.
One typical method is \textit{GroupDRO} given in~\citep{sagawa2019DRO}, of which the objective is:
\begin{small}
\begin{equation}
\min_{f_{\theta}:\mathcal{X}\rightarrow \mathcal{Y}} \max_{q\in \Delta_g}\sum_{e\in \mathcal{E}_{train}}q_e\cdot \mathcal{R}(f_{\theta};e), \tag{GroupDRO}
\end{equation}
\end{small}where $g=\lvert \mathcal{E}_{train}\rvert$ is the numbers of accessible training environments. 
Similarly, we define that $\mathcal{L}_{GroupDRO}(\theta; \mathcal{E}_{train})=\max_{q\in \Delta_g}\sum_{e\in \mathcal{E}_{train}}q_e\cdot \mathcal{R}(f_{\theta};e)$ in this paper.

\subsection{Personalized invariance}
\label{per-invariance}

\begin{table}[t]
  \caption{The coverage of the related works.}
  \label{table:related}
  \centering
  \small
  \resizebox{\linewidth}{!}{
\setlength{\tabcolsep}{1.5mm}{
  \begin{tabular}{lccc}
    \toprule[1.5px]
    Methods & Collaboration & Personalization  & OOD Generalization \\
    \midrule
    Federated learning, e.g.,~\citep{mcmahan17} & \Checkmark & \XSolidBrush & \XSolidBrush \\
    Personalized FL, e.g.,~\citep{dinh20pFedMe,DBLP:conf/nips/0001MO20,li2021ditto,cheng2021fine} & \Checkmark    & \Checkmark & \XSolidBrush \\
    Invariant learning, e.g.,~\citep{arjovsky2019IRM,sagawa2019DRO,creager2021EIIL} & \XSolidBrush    & \XSolidBrush    & \Checkmark \\
    \midrule
    \rowcolor{mygray}PerInvFL (Ours) & \Checkmark & \Checkmark & \Checkmark \\
    \bottomrule[1.5px]
  \end{tabular}
  }
  }
  \vspace{-0.4cm}
\end{table}

In order to study the OOD generalization problem under Non-IID federated setting, we first dissect the data heterogeneity from two orthogonal perspectives and categorize the related works in Table~\ref{table:related} according to the coverage of their methods.

\textbf{Client-level heterogeneity.} The distributions of training datasets are usually heterogeneous across the clients. 
Since the local data on every client is scarce and of limited diversity, the clients in FL framework need to collaborate to mine the common/generalization knowledge with some shared/global objective. However, the client-level heterogeneity can result in the discrepancy between the optimas of local objective and global objective. In other words, the client-independent representation (i.e., global knowledge) can be biased for some participating clients. Therefore, preserving the personalization information is of great importance for dealing with the client-level data heterogeneity.  

\textbf{Context-level heterogeneity.} In addition to the heterogeneity among the training datasets, the heterogeneity between the training and testing dataset is also worthy of attentions. In FL, the local training and testing dataset on each client may be generated in distinct contexts/environments\footnote{In this paper, environment, context and domain are used equivalently.}. For example, the training image samples on a client are mainly captured from the local cameras, while the testing images may come from the web and with different styles. From the perspective of context, the target learning models must be equipped with OOD generalization so that they can perform both well on unknown and distinct contexts. 

As discussed above, we can conclude that solving the OOD problem under Non-IID federated setting needs to discover the invariant 
representation without dropping the important personalized information. When we consider to implement the distributed version of invariant learning, such as \textit{IRM} and \textit{GroupDRO}, each client can be regarded as an environment. In this way, both the context-specific (i.e., spurious) representation and the client-related (personalized) representation will be treated as spurious representations. Only the common invariant representation across the clients is extracted and utilized to elicit the invariant predictors. Consequently, the important personalized representation is dropped.

\begin{assumption}
(\textbf{Heterogeneity of invariance}) For each client $i$, there exists a set of invariant representations $\Phi_{i}$ that satisfies the invariance property on the local set of environments $\mathcal{E}^{i}_{all}$ as defined in \eqref{invariance}. 
There are $K=pN$ clients ($[N_K]\subset [N]$) on each of which the corresponding set of invariant representations $\Phi_{j} (j\in[N_K])$ contain some representations $Z_{j}$ satisfying: 
\begin{enumerate}[label=(\roman*)]
\item \label{subass:1}\textbf{Heterogeneity:} $Z_{j}$ is a subset of $\Phi_{j}$ and not contained in the intersection $\mathop{\bigcap}_{i\ne j, i\in [N]}\Phi_{i}$,
\item \label{subass:2}\textbf{Informativeness:} $\max_{z\subset Z_{j}}I(Y; z) = \delta > 0$, where $I(\cdot;\cdot)$ measures the mutual information (\textbf{MI}) between two random variables.
\end{enumerate}
\label{ass:1}
\end{assumption}

The heterogeneity condition in Assumption~\ref{ass:1} is consistent with the Non-IID assumption in FL, and the informativeness assumption claims the existence of meaningful personalization. If we define \textit{global invariant} representation as $\Phi_{g}=\mathop{\bigcap}_{i\in [N]}\Phi_{i}$ and \textit{personalized invariant} representation as $\Phi_{i}$, $i\in [N]$. In the prediction task, the optimal global and personalized invariant representations are given by $\Phi_{i}^{\star}=\arg\max_{S\subset \Phi_i} I(Y; S)$ and $\Phi_{g}^{\star}=\arg\max_{S\subset \Phi_{g}} I(Y; S)$, respectively.

\begin{theorem}
If the Assumption~\ref{ass:1} holds in FL, the proposed \textit{personalized invariant} representations can be constantly more informative than the global invariant representations obtained by the distributed implementation of existing invariant learning, for the prediction performance. That is,
\begin{small}
\begin{equation}
\frac{1}{N}\sum_{i=1}^N I(Y; \Phi_i^{\star}) \geq \frac{1}{N}\sum_{i=1}^N I(Y; \Phi_g^{\star}) + p{\delta},
\end{equation}
\end{small}\\
where $0<p<1$ is a constant and $\delta$ is a positive constant that is independent of $N$.
\label{theorem:1}
\end{theorem}

Theorem~\ref{theorem:1} demonstrates the importance of the proposed personalized invariance in improving the average prediction performance for OOD problems under Non-IID federated setting.
The detailed proof of Theorem~\ref{theorem:1} is provided in the Appendix.

\section{Dual-regularization constrained learning framework}
\label{framework}
Adhering to the prevalent setup of invariant learning, we suppose there are totally $K$ training contexts in the federated learning system. On client $i$, the training data $(X, y)\in D_i$ is generated from $K_i$ contexts while the testing data comes from unknown contexts. That is, $\mathcal{E}_{train}^i = \{e_j^i| j\in S_i\}$, where $S_i$ is a subset of $[K]$ and the size of $S_i$ equals $K_i$.
\subsection{Algorithm Design}
\label{algo}

In ideal circumstances, we can elicit the personalized invariant predictor for each client $i$ $(i\in[N])$ using $\theta_i^{\star}\in \argmin_{\theta_i} \mathcal{L}_{INV}(\theta_i; \mathcal{E}_{train}^i)$, where $\mathcal{L}_{INV}\in \{\mathcal{L}_{IRM}, \mathcal{L}_{GroupDRO}\}$. We refer to `ideal' circumstances as the scenarios where the following two conditions are met: \textbf{1)} the number of training contexts on every client is sufficient for deriving the invariant predictor, \textbf{2)} the data samples in every training context are of sufficient diversity. Unfortunately, both of the above two conditions can hardly be satisfied in FL setting since the local data is usually scarce~\citep{mcmahan17}. On the one hand, the latest efforts prove that the minimizer of $\mathcal{L}_{IRM}$ will necessarily utilize the non-invariant features and therefore cannot universally generalize to unknown testing contexts when the number of training contexts is not sufficient~\citep{rosenfeld2021IRMrisks}. On the other hand, the limited data diversity in each training contexts will make the obtained model easily trapped by overfitting. 

To handle the challenges of insufficient training contexts and limited data diversity in each available context . In this paper, we elaborate a dual-regularization constrained optimization framework to handle these challenges of OOD generalization problems in federated scenarios. The overall objective on client $i (i\in [N])$ is: 
\begin{small}
\begin{align}
&\min_{\theta_i} \mathcal{L}_{INV}(\theta_i; \mathcal{E}_{train}^i) + \beta{\lVert \theta_i - {\nu}^{\star}\rVert}^2 \label{ob:per-inv}\\
&\text{s.t.}\quad {\nu}^{\star}\in\argmin_{\nu}\mathcal{L}_{INV}(\nu; \mathcal{E}^D) \label{ob:glob-inv}
\end{align}
\end{small}\\
where $\mathcal{E^D}=\{D_i|i=1,2,...,N\}$ and $\mathcal{L}_{INV} \in \{\mathcal{L}_{IRM}, \mathcal{L}_{GroupDRO}\}.$ Different from the popular single-regularized (i.e., global-regularized) personalized federated learning schemes~\citep{dinh20pFedMe,hanzely20mix,deng20peradap,hanzely20lower,li2021ditto}, the objective on each client is constrained by two regularizations. One is the `local invariance' expressed by the first term $\mathcal{L}_{INV}(\theta_i; \mathcal{E}_{train}^i)$ and nother is the `global invariance' represented by the second term ${\lVert \theta_i - \nu^{\star}\rVert}^2$. The `local invariance' is necessary for discovering the personalized invariant predictor, while the `global invariance' is adopted to deal with the challenges of insufficient training contexts and limited data diversity in each context. 
Without loss of generality, we focus on studying the case where $\mathcal{L}_{INV}=\mathcal{L}_{IRM}$ in this paper. Therefore, the detailed objective on client $i$ can be written as
\begin{small}
\begin{align}
&\min_{\theta_i} \sum_{j=1}^{K_i}\mathcal{R}(f_{\theta_i};e_j^i) + \lambda\cdot{\lVert \nabla_{\bar{\omega}|\bar{\omega}=1.0}\mathcal{R}(\bar{\omega}\cdot f_{\theta_i};e_j^i)\rVert}^2 + \beta{\lVert \theta_i - {\nu}^{\star}\rVert}^2 \label{ob:per-irm}\\
&\text{s.t.}\quad {\nu}^{\star}\in\argmin_{\nu}\sum_{i=1}^N \mathcal{R}(f_{\nu};D_i) + \lambda\cdot{\lVert \nabla_{\bar{\omega}|\bar{\omega}=1.0}\mathcal{R}(\bar{\omega}\cdot f_{\nu};D_i)\rVert}^2 \label{ob:glob-irm}
\end{align}
\end{small}
On each client, the outer problem in objective (\ref{ob:per-inv}) can be solved locally with the guidance of the global model $\nu^{\star}$ downloaded from the server. In order to solve the inner problem in a collaborative manner, the objective (\ref{ob:glob-inv}) needs to be decomposed into $N$ subproblems that can be individually and parallelly solved at the local clients.
In fact, the objective of \textit{IRM} ($\mathcal{L}_{IRM}(\nu;\mathcal{E}_D)$) can be naturally decomposed into $N$ subproblems: $\mathcal{L}_{IRM}^i=\mathcal{R}(f_{\nu};D_i) + \lambda \cdot{\lVert\nabla_{\bar{w}|\bar{w}=1.0}\mathcal{R}(\bar{w}\cdot f_{\nu};D_i)\rVert}^2, i\in[N]$. As regard to \textit{GroupDRO}, the objective $\mathcal{L}_{GroupDRO}(\nu;\mathcal{E_D})$ can be decomposed into $\mathcal{L}_{GroupDRO}^i = q_i\cdot \mathcal{R}(f_{\nu};D_i), i\in [N]$. The weight $q_i$ can be updated on the server using the uploaded $\mathcal{R}(f_{\nu};D_i), i \in [N]$ following the update law in~\citep{sagawa2019DRO}.
Our algorithm is shown in Algorithm~\ref{algo:perinvfl}. 

\begin{algorithm}[htb]
\begin{small}
\caption{\textit{PerInvFL}: Personalized Invariant Federated Learning}
\label{algo:perinvfl}
\textbf{Input}: $T, R, S, \beta, \eta, \gamma, \alpha$. 
\end{small}
\begin{small}
\begin{algorithmic}[1]
\State Initialize the models $\nu^0$, $\{\theta_i^0|i\in [N]\}$.
  \For {$t=0$ to $T-1$}
    \State Server sends the global model ($\nu^t$) to the participating clients.
    \For {local device $i=1$ to $N$ in parallel}
      \State Initialization: $\nu_{i,0}^t \gets \nu^t$.
      \For {$r=0$ to $R-1$}
        \For {$s=0$ to $S-1$}
            \State Update the personalized model: $\theta_i^{s+1} \gets \theta_i^s - \eta \big(\nabla \mathcal{L}_{INV}(\theta_i^s;\mathcal{E}_{train}^i) + 2\beta(\theta_i^s - \nu_{i,r}^t)\big)$.
        \EndFor
        \State Local update for global model: $\nu_{i,r+1}^t \gets \nu_{i,r}^t - \gamma\nabla \mathcal{L}_{INV}(\nu_{i,r}^t;D_i)$.
      \EndFor
    \EndFor
    \State Global aggregation: $\nu^{t+1} \gets \nu^t - \alpha\big(\nu^t - \frac{1}{N}\sum_{i\in [N]}\nu_{i,R}^t\big)$.
  \EndFor
  \State {\textbf{return} the personalized models $\{\theta_i^S | i\in[N]\}$ and global model $\nu^T$.}
\end{algorithmic}
\end{small}
\end{algorithm}
\vspace{-0.3cm}

\subsection{Theoretical analysis}
\label{theory}
In this section, we will provie detailed theoretical analysis on convergence rate and out-of-distribution generalization guarantee of the proposed algorithm, and discuss how to improve the practical performance of our method based on the theoretical outcomes. For simplicity, we define $\mathcal{L}(\nu):=\frac{1}{N}\mathcal{L}_{IRM}(\nu;\mathcal{E}^D)$ in this section. It's mentioned that the penalty term of \textit{IRM} is generally non-convex in~\citep{arjovsky2019IRM}. Therefore, we figure out the convergence rate of our algorithm under the non-convex and smooth case. 

\noindent\textbf{Assumption 2} \textbf{(Bounded variance)} The variance of local gradients to the aggregated average is upper bounded by
$\frac{1}{N}\sum_{i=1}^{N}{\lVert \nabla\mathcal{L}_{IRM}^i(\nu;D_i) - \nabla\mathcal{L}_{IRM}(\nu;\mathcal{E}^D) \rVert}^2 \leq \delta_L^2$, where $\delta_L$ is a finite constant.

\begin{theorem}
\textbf{(Convergence)} Suppose the IRM loss function $\mathcal{L}_{IRM}(\theta;e_j^i)$ is $L_F$-smooth, $\forall$ $i$, $j$, and Assumption 2 holds. If $\gamma \leq \frac{\gamma_0}{\alpha R}$ $\forall$ $\alpha\leq 1$, $R\geq 1$ and $\gamma_0:=\frac{\alpha}{\sqrt{32(R+3)}L_F}$. Then we have
\begin{enumerate}
\item The convergence rate of the global model is given by: \\
	  \begin{small}
	  \begin{equation*}
	  \mathbb{E}\big[{\lVert \nabla\mathcal{L}(\nu_{t^{\star}}) \rVert}^2 \big] \leq \mathcal{O}\big( \mathbb{E}\big[{\lVert \nabla\mathcal{L}(\nu_{t^{\star}}) \rVert}^2\big] \big) := \mathcal{O}\Big( \frac{\Delta_\mathcal{L}R^{\frac{1}{2}}L_F}{\beta T} + \frac{{(\Delta_\mathcal{L}L_F)}^{\frac{3}{4}} {(R\delta_L^2)}^{\frac{1}{4}} }{\alpha^{\frac{1}{2}} T^{\frac{3}{4}}} + \frac{{(\Delta_\mathcal{L} \delta_L L_F)}^{\frac{2}{3}}}{\alpha^{\frac{2}{3}} T^{\frac{2}{3}}} \Big).
	  \end{equation*}
	  \end{small}
\item And, the convergence rate of the personalized models is given by: \\
	  \begin{small}
	  \begin{equation*}
	  \frac{1}{N}\sum_{i=1}^N\mathbb{E}\big[{\lVert \theta_i^{S}(\nu_{t^{\star}}) - \nu_{t^{\star}} \rVert}^2\big] \leq \mathcal{O}\Big( \mathbb{E}\big[{\lVert \nabla\mathcal{L}(\nu_{t^{\star}}) \rVert}^2\big] \Big) + C,
	  \end{equation*}
	  \end{small}
\end{enumerate} where $\Delta_\mathcal{L}:=\mathcal{L}(\nu_0) - \mathcal{L}(\nu_T)$, $C$ is a finite positive constant, and $t^{\star}$ is uniformly sampled from $\{0, 1, ..., T-1\}$. The constant $C$ exists since the number of local iterations $S$ is finite. 
\label{theo:conver}
\end{theorem}

\noindent\textbf{Corollary 1} Theorem~\ref{theo:conver} proves that the proposed algorithm achieves a convergence rate of $\mathcal{O}(1/T^{2/3})$ under the non-convex and smooth case. The detailed proof is provided in the Appendix.

\begin{theorem}
\textbf{(Out-of-distribution Generalization)} As discussed in Section~\ref{formulation}, the learning model can be written as $f=\Phi\cdot \omega$. Suppose Assumption~\ref{ass:1} is satisfied. Let the global invariant featurizer $\Phi_g^{\star}\in \mathbb{R}^{d\times d}$ have rank $\tau>0$, and the personalized invariant featurizer $\Phi_i^{\star}$ have rank $\tau + q_i$ where $q_i>0$ $(i\in[N])$ and $d>\max_{i\in[N]} q_i + \tau$. Denote the minimizers of our designed objective by $\{f_{\hat{\theta}_i}=\hat{\Phi}_i \cdot \hat{\omega}_i\}$. Then, if there are at least $d-\tau+\frac{d}{\tau}$ local datasets in $\mathcal{E}^D$ lying in a linear general position of degree $\tau$, 
the following holds on each client $i$ $(i\in[N])$:
\begin{enumerate}
	\item When there are at least $d-(\tau+q_i)+\frac{d}{\tau+q_i}$ training contexts in $\mathcal{E}_{train}^i$ lying in a linear general position of degree $\tau+q_i$, with $\beta$ appropriately chosen, we can guarantee that $\mathbb{E}[Y|\hat{\Phi}_i(X),e] = \mathbb{E}[Y|\hat{\Phi}_i(X),e^{\prime}]$, for all $e, e^{\prime}\in \mathcal{E}_{all}^i$, and the test error
	$\mathbb{E}_{(X,Y)\sim \mathbb{P}^{e}}[\ell(f_{\hat{\theta}_i}(X),Y)] = \min_{\omega, \Phi_i^{\star}}\mathbb{E}_{(X,Y)\sim \mathbb{P}^{e}}[\ell(\omega(\Phi_i^{\star}(X)),Y)]$ for any $e \in \mathcal{E}_{all}^i$.
	\item Otherwise, with $\beta$ appropriately chosen, we can guarantee that $\mathbb{E}[Y|\hat{\Phi}_i(X),e] = \mathbb{E}[Y|\hat{\Phi}_i(X),e^{\prime}], for\text{ }all\text{ }e, e^{\prime}\in \mathcal{E}_{all}^i$.
\end{enumerate}
\label{theorem:2}
\end{theorem}

\noindent\textbf{Remark 1} Theorem~\ref{theorem:2} provides the personalized models with performance guarantee on any unseen contexts, even if both challenges of insufficient training contexts and limited data diversity in each available context exist. The detailed proof of Theorem~\ref{theorem:2} can be found in the Appendix. In particular, we find that choosing a appropriate value for the balancing weight $\beta$ is significant for guaranteeing the out-of-distribution performance on each client. Empirically, the less training contexts a client has, the smaller the value of $\beta$ should be. Therefore, personalized $\beta$ can be adopted on heterogeneous clients by themselves, according to the number of available training contexts.

\section{Experiments}
\label{experiments}

\subsection{Experimental Setup}
\paragraph{Datasets} 
We evaluate the effectiveness of our method on three datasets that are frequently adopted in the related literature:
\textbf{1)} \textbf{Rotated CMNIST (RC-MNIST)}, a variant of CMNIST~\citep{arjovsky2019IRM}. 
Firstly, We construct the CMNIST dataset according to the same law as in~\citep{arjovsky2019IRM}:
each image having label 0 is colored green with probability $p^e$ and colored red with probability $1-p^e$. 
In contrary, each image having label 1 is colored red with probability $p^e$ and colored green with probability $1-p^e$. 
In the training and testing contexts, we set $p^e=p^e_{train}$ and $p^e=p^e_{test}$ to generate the data respectively.
On the basis of CMNIST, we add rotation property to simulate the personal writing style and regard the rotation property as useful personalized information. To distribute the data to the clients in a Non-IID scheme, we construct four training contexts with $p^e_{train}=0.95, 0.90, 0.85, 0.80$ and assign each context to a client as training data. For the testing data, we constrcut a out-of-distribution context with $p^e_{test}=0.10$ for every client. The data used for constructing the four training/testing contexts is randomly sampled from the train-set/test-set of MNIST~\citep{lecun1998MNIST} without replacement.
In addition, all images (from both training and testing sets) in the four clients are rotated by $0^{\circ}$, $90^{\circ}$, $180^{\circ}$ and $270^{\circ}$ respectively.
\textbf{2)} \textbf{Rotated Colored Fashion-MNIST (RC-FMNIST)}~\citep{ahuja2020IRMgames}, which is constructed using the same strategy as RC-MNIST, except that the original images come from FashionMNIST dataset. 
\textbf{3)} \textbf{WaterBird}~\citep{sagawa2019DRO}, which is constructed by placing the waterbird photographs onto the water background with probability $p^e$ and onto water background with probability $1-p^e$. In the meanwhile, the landbird photographs are placed onto the land background with probability $p^e$ and onto water background with probability $1-p^e$. In this paper, we construct four training contexts with $p^e_{train}=0.95, 0.90, 0.85, 0.80$ and the testing contexts with $p^e_{test}=0.10$. To distribute the data in a Non-IID manner, we construct the four training contexts using the bird photograghs of disjoint bird classes from the bird dataset and background photographs randomly selected from the background dataset without replacement. Each training context is assigned to a client as training data. For the testing data in each client, we construct a testing contexts with probability $p^e_{test}=0.10$, using the bird photographs of same bird classes (but distinct instances) as the corresponding training context and the background photographs are randomly selected from the background dataset. 
\paragraph{Model selection and Competitors}
For RC-MNIST and RC-FMNIST dataset, we adopt a deep neural network with two hidden layers as the feature extractor and an subsequent fully-connected layer as the classifier. In the evaluation on WaterBird dataset, we adopt the ResNet-18~\citep{he2016resnet} as learning model. Similarly, the part before the last fully-connected layer works as the feature extractor and the last fully-connected layer works as the classifier.
We compare our method (\textit{PerInvFL}) with six state-of-the-art algorithms in federated learning and invariant learning, including one traditional federated learning method (\textit{FedAvg}~\citep{mcmahan17}), three personalized learning methods (\textit{pFedMe}~\citep{dinh20pFedMe}, \textit{Ditto}~\citep{li2021ditto}, and \textit{FTFA}~\citep{cheng2021fine}), and two invariant learning methods (\textit{IRM}~\citep{arjovsky2019IRM} and \textit{GroupDRO}~\citep{sagawa2019DRO}).


\begin{table}[ht]
  \caption{Results on Rotated-Colored MNIST dataset}
  \label{table:res_rcmnist}
  \centering
  \resizebox{\textwidth}{16mm}{
  \begin{tabular}{lcccccc}
    \toprule
               & OOD case 1 & OOD case 2 & OOD case 3 & OOD case 4 & OOD case 5 & Average \\
    $p^e_{test}$& 0.10      & 0.20       & 0.30       & 0.40       & 0.50       &         \\            
    \midrule
    FedAvg     & $20.77 (\pm0.0)$ & $29.66 (\pm0.0)$ & $37.74 (\pm0.0)$ & $45.15 (\pm0.0)$ & $54.12 (\pm0.0)$ & $37.49 (\pm11.6)$   \\
    Ditto      & $23.62 (\pm0.0)$ & $31.85 (\pm0.0)$ & $39.57 (\pm0.2)$ & $46.40 (\pm0.0)$ & $\mathbf{54.92 (\pm0.0)}$ & $39.27 (\pm10.9)$  \\
    FTFA       & $19.86 (\pm0.1)$ & $28.87 (\pm0.1)$ & $37.37 (\pm0.0)$ & $43.75 (\pm0.1)$ & $53.69 (\pm0.0)$ & $36.71 (\pm11.7)$ \\
    pFedMe     & $30.28 (\pm0.0)$ & $38.98 (\pm0.0)$ & $40.05 (\pm0.0)$ & $44.48 (\pm0.0)$ & $49.67 (\pm0.0)$ & $40.69 (\pm6.43)$   \\
    GroupDRO   & $29.32 (\pm0.1)$ & $35.68 (\pm0.0)$ & $41.50 (\pm0.0)$ & $46.93 (\pm0.0)$ & $53.77 (\pm0.2)$ & $41.44 (\pm8.51)$ \\
    IRM        & $51.48 (\pm0.0)$ & $51.70 (\pm0.1)$ & $52.17 (\pm0.1)$ & $\mathbf{51.53 (\pm0.0)}$ & $51.18 (\pm0.0)$ & $51.61 (\pm0.33)$  \\
    \midrule
    PerInvFL   & $\mathbf{53.46 (\pm0.2)}$ & $\mathbf{52.75 (\pm0.1)}$ & $\mathbf{53.63 (\pm0.1)}$ & $51.41 (\pm0.3)$ & $52.73 (\pm0.1)$ & $\mathbf{52.80 (\pm 0.78)}$   \\
    \bottomrule
  \end{tabular}}
  \vspace{-0.4cm}
\end{table}

\begin{table}[ht]
  \caption{Results on Rotated Colored-FMNIST dataset}
  \label{table:res_cfmnist}
  \centering
  \resizebox{\textwidth}{16mm}{
  \begin{tabular}{lcccccc}
    \toprule
               & OOD case 1 & OOD case 2 & OOD case 3 & OOD case 4 & OOD case 5 & Average \\
    $p^e_{test}$& 0.10      & 0.20       & 0.30       & 0.40       & 0.50       &         \\            
    \midrule
    FedAvg     & $10.70 (\pm0.0)$ & $20.80 (\pm0.0)$ & $29.85 (\pm0.0)$ & $41.84 (\pm0.0)$ & $50.54 (\pm0.0)$ & $30.75 (\pm14.3)$ \\
    Ditto      & $16.35 (\pm0.0)$ & $25.75 (\pm0.1)$ & $34.11 (\pm0.0)$ & $45.09 (\pm0.0)$ & $52.59 (\pm0.0)$ & $34.78 (\pm13.0)$ \\
    FTFA       & $17.32 (\pm0.1)$ & $26.48 (\pm0.0)$ & $34.36 (\pm0.0)$ & $45.43 (\pm0.0)$ & $53.03 (\pm0.0)$ & $35.32 (\pm12.8)$ \\
    pFedMe     & $20.32 (\pm0.0)$ & $27.40 (\pm0.1)$ & $34.40 (\pm0.0)$ & $42.85 (\pm0.0)$ & $48.63 (\pm0.0)$ & $34.72 (\pm10.2)$ \\
    GroupDRO   & $30.13 (\pm0.0)$ & $36.00 (\pm0.0)$ & $41.33 (\pm0.0)$ & $47.55 (\pm0.1)$ & $52.59 (\pm0.0)$ & $41.52 (\pm7.99)$ \\
    IRM        & $47.35 (\pm0.1)$ & $48.99 (\pm0.1)$ & $50.53 (\pm0.1)$ & $51.57 (\pm0.0)$ & $52.66 (\pm0.1)$ & $50.22 (\pm1.88)$ \\
    \midrule
    PerInvFL   & $\mathbf{51.71 (\pm0.2)}$ & $\mathbf{52.10 (\pm0.1)}$ & $\mathbf{50.77 (\pm0.2)}$ & $\mathbf{53.12 (\pm0.1)}$ & $\mathbf{53.64 (\pm0.1)}$ & $\mathbf{52.27 (\pm 1.02)}$ \\
    \bottomrule
  \end{tabular}}
\end{table}

\begin{table}[ht]
  \caption{Results on WaterBird dataset. }
  \label{table:res_waterbird}
  \centering
  \resizebox{\textwidth}{16mm}{
  \begin{tabular}{lcccccc}
    \toprule
               & OOD case 1 & OOD case 2 & OOD case 3 & OOD case 4 & OOD case 5 & Average \\
    $p^e_{test}$& 0.10      & 0.20       & 0.30       & 0.40       & 0.50       &         \\            
    \midrule
    FedAvg     & $51.33 (\pm0.77)$ & $60.08 (\pm0.47)$ & $57.75 (\pm0.21)$ & $62.50 (\pm0.61)$ & $59.25 (\pm1.62)$ & $58.18 (\pm3.76)$  \\
    Ditto      & $51.17 (\pm1.31)$ & $60.58 (\pm0.68)$ & $58.74 (\pm0.53)$ & $63.50 (\pm0.74)$ & $60.17 (\pm0.66)$ & $58.83 (\pm4.13)$ \\
    FTFA       & $51.67 (\pm0.69)$ & $61.50 (\pm0.71)$ & $59.00 (\pm0.43)$ & $63.00 (\pm0.83)$ & $60.25 (\pm0.94)$ & $59.08 (\pm3.94)$ \\
    pFedMe     & $52.17 (\pm0.12)$ & $56.42 (\pm0.12)$ & $58.75 (\pm0.01)$ & $65.00 (\pm0.03)$ & $62.25 (\pm0.00)$ & $58.92 (\pm4.47)$ \\
    GroupDRO   & $61.33 (\pm0.31)$ & $63.25 (\pm0.35)$ & $65.75 (\pm1.03)$ & $69.05 (\pm1.08)$ & $62.58 (\pm0.31)$ & $64.39 (\pm2.74)$ \\
    IRM        & $62.75 (\pm0.41)$ & $62.75 (\pm0.00)$ & $65.25 (\pm0.23)$ & $66.42 (\pm0.12)$ & $63.42 (\pm0.59)$ & $64.12 (\pm1.47)$ \\
    \midrule
    PerInvFL   & $\mathbf{71.17 (\pm0.44)}$ & $\mathbf{74.15 (\pm0.37)}$ & $\mathbf{73.67 (\pm0.21)}$ & $\mathbf{75.13 (\pm0.11)}$ & $\mathbf{75.83 (\pm0.34)}$ & $\mathbf{73.99 (\pm 1.60)}$ \\
    \bottomrule
  \end{tabular}}
  \vspace{-0.4cm}
\end{table}

\subsection{Experimental Results}
In order to comprehensively compare the performance of our method with that of the baselines, we conduct evaluation on five out-of-distribution (OOD) testing cases for each dataset, including $p^e_{test} = 0.10, 0.20, 0.30, 0.40$ and $0.50$. All hyperparameters of the algorithms are tuned optimal, and more details are provided in the Appendix. We run each algorithm three times with different random seeds to record the mean and standard deviation of the test accuracy in every OOD case. 

\textbf{Performance Comparison.}
The overall results on RC-MNIST, RC-FMNIST and WaterBird are displayed in Table~\ref{table:res_rcmnist}, Table~\ref{table:res_cfmnist} and Table~\ref{table:res_waterbird}, respectively. In each table, the last column gives the mean and standard deviation of the results in the five OOD cases. From the results shown in the three tables, we can find that 1) our method \textit{PerInvFL} can constantly outperform the baseline methods on both worst-case accuracy and average accuracy for diverse datasets. In particular, our method achieves an average accuracy of $73.99\%$ which is about $\mathbf{9\%}$ \textbf{higher} than the second highest one on WaterBird dataset. 2) As shown in the last columns, the performance deviation of our method is much smaller than the existing federated learning methods and approximate to the invariant learning approach \textit{IRM}, demonstrating that our method can effectively obtain the invariant predictor and hence achieve consistent performance in different out-of-distribution cases. These findings verify that the proposed method can effecrively extract the invarinat features and concurrently exploit the personalized information to improve the performance on out-of-distribution testing data.

\begin{table}[!h]
  \caption{The effect of the elaborated dual-regularization}
  \label{table:regs}
  \centering
  \begin{small}
  \setlength\tabcolsep{4pt}
  \begin{tabular}{lccccccccccc}
    \toprule
     & \multicolumn{5}{c}{Rotated-CMNIST} & & \multicolumn{5}{c}{WaterBird} \\
    \cmidrule(r){2-6}
    \cmidrule(r){8-12}
    $p^e_{test}$ & 0.10 & 0.20 & 0.30 & 0.40 & 0.50 & & 0.10 & 0.20 & 0.30 & 0.40 & 0.50 \\            
    \midrule
    IRM        & $51.48$ & $51.70$ & $52.17$ & $\mathbf{51.53}$ & $51.18$ & & $62.75$ & $62.75$ & $65.25$ & $66.42$ & $63.42$  \\
    IRM-$L2$   & $31.19$ & $29.35$ & $36.53$ & $43.06$ & $51.83$ & & $56.83$ & $62.75$ & $58.75$ & $66.00$ & $59.42$ \\
    IRM-FT     & $21.43$ & $22.37$ & $32.42$ & $40.32$ & $50.96$ & & $55.67$ & $62.50$ & $65.05$ & $66.00$ & $63.75$ \\
    \midrule
    PerInvFL   & $\mathbf{53.46}$ & $\mathbf{52.75}$ & $\mathbf{53.63}$ & $51.41$ & $\mathbf{52.73}$ & & $\mathbf{71.17}$ & $\mathbf{74.15}$ & $\mathbf{73.67}$ & $\mathbf{75.13}$ & $\mathbf{75.83}$ \\
    \bottomrule
  \end{tabular}
  \end{small}
\end{table}
\textbf{Effect of dual-regularized optimization.}
As explained in section~\ref{framework}, the elaborated dual-regularized optimization is formulated as the objective~(\ref{ob:per-inv}). To demonstrate the effect of the designed dual-regularized franework, we replace the dual-regularized optimization with some prevalent single-regularized methods that are usually adopted in the state-of-the-art personalized federated learning, and keep the rest part of the overall objective (shown in equation~(\ref{ob:per-inv}) and~(\ref{ob:glob-inv})) unchanged. Concretely, we implement two typical personalization skills with the global model being achieved by distributional \textit{IRM} that we introduce in section~\ref{algo}. One is the $L2$-norm regularizer which is widely used in PFL~\citep{dinh20pFedMe,hanzely20mix,hanzely20lower,li2021ditto}, and we call this implementation \textit{IRM-$L2$}. Another one is the local-finetune skill which is proved simple and effective in~\citep{cheng2021fine}. In particular, the local-finetune skill can be viewed as a `none' regularizer. We name the implementation with local-finetune skill as \textit{IRM-FT}. 

The comparison of results is shown in Table~\ref{table:regs}. We can find that combining the existing personalization terms with \textit{IRM} cannot improve the performance of OOD problem under federated setting. On the contrary, the widely adopted personalization term ($L2$-norm and local-finetune) can even degrade the OOD performance, compared with the distributional version of \textit{IRM}. The underlying reason is that the local objective with these personalization terms is to minimize the expected loss as long as the target model is not too different from the global model. However, when the realtionship with the global model is satisfied, minimizing the expected loss can readily make the trained model pick up the spurious correlations and hence decrease the performance on OOD testing data. By comparison, the designed dual-regularized optimization in our algorithm contains two constraints: 1) being close/similar to the `global invariance' and 2) satisfying the `local invariance' to exclude the spurious features. Therefore, our method can effectively exploit the personalized information and avoid `picking up' the spurious information concurrently, so as to improve the OOD performance. 

\section{Conclusion}
\label{conclusion}
In this paper, we are the first to investigate the out-of-distribution problem under the Non-IID federated setting. Atfer formally analyzing the challenges, we propose the novel concept \textit{personalized invariance} which can improve the model performance, by preserving the important personalized information and meanwhile eliminating the spurious correlations. To explore the optimal personalized invariance, we propose a objective: dual-regularization constrained optimization and design a practical algorithm \textit{PerInvFL} to solve it. Both the theoretical and experimental results demonstrate the superiority of our method over the state-of-the-art federated learning and invariant learning methods. 


\bibliography{pFedInv-references}
\bibliographystyle{plainnat}

\appendix
\section{Appendix}
\label{appendix}
In this appendix, we provide more details that are simplified in the main text duo to page limitation, including the complemte proofs, more details on experimental setup, and some further discussions.


\subsection{Proof of Theorem 1}
\label{proof:T1}
\textbf{Theorem 1}
If the Assumption 1 holds in FL, the proposed \textit{personalized invariant} representations can be constantly more informative than the global invariant representations obtained by the distributed implementation of existing invariant learning, for the prediction performance. That is,
\begin{equation}
\frac{1}{N}\sum_{i=1}^N I(Y; \Phi_i^{\star}) \geq \frac{1}{N}\sum_{i=1}^N I(Y; \Phi_g^{\star}) + p{\delta},
\end{equation} where $0<p<1$ is a constant and $\delta$ is a positive constant that is independent of $N$.

\textit{Proof:}
According to the definition, there are $N$ clients in the federated learning system. The set of all possible environments on each client $i$ is given by $\mathcal{E}_{all}^i$, $i \in [N]$. Then, the set of all possible environments on all the clients is the union:
\begin{equation}
\label{equ:envs}
\mathcal{E}_{all} = \mathop{\bigcup}_{i\in[N]}\mathcal{E}_{all}^i
\end{equation}
Because of the data heterogeneity across the clients, we have $\mathcal{E}_{all}^i \ne \mathcal{E}_{all}$, i.e., $\mathcal{E}_{all}^i \subset \mathcal{E}_{all}$, $\forall i \in [N]$.
For every invariant feature $\phi_g\in \Phi_g$ across all the environments $\mathcal{E}_{all}$, we can get
\begin{equation}
\phi_g \in \Phi_i, \forall i \in [N].
\end{equation}
But, the converse is not true, due to the result in equation~\ref{equ:envs}. Therefore, we have
\begin{equation}
\Phi_g \subseteq \Phi_i, \forall i \in [N]
\end{equation}
When the \textbf{Heterogeneity} condition in Assumption 1 is satisfied, we can conclude that
\begin{equation}
\begin{cases}
\Phi_g \subset \Phi_j, \forall j \in [N_K], \\
\Phi_g = \Phi_j, \forall j \notin [N_K] \text{ and } j\in[N].
\end{cases}
\end{equation}
When the \textbf{Informativeness} condition in Assumption 1 is satisfied, we can derive the average mutual-information among the clients by
\begin{align*}
\frac{1}{N}\sum_{i=1}^{N}I(Y;\Phi_i^{\star}) &= \frac{1}{N}\sum_{i=1}^{N}\max_{S_i\subseteq \Phi_i} I(Y;S_i) \\
&= \frac{1}{N}\sum_{i\in[N_K]}\max_{S_i\subseteq \Phi_i} I(Y;S_i) + \frac{1}{N}\sum_{i\notin[N_K]}\max_{S_i\subseteq \Phi_i} I(Y;S_i) \\
&\geq \frac{1}{N}\sum_{i\in[N_K]}\big\{\max_{S_{i,1}\subseteq \Phi_g} I(Y;S_{i,1}) + \max_{S_{i,2}\subseteq Z_i} I(Y;S_{i,2})\big\} + \frac{1}{N}\sum_{i\notin[N_K]}\max_{S_i\subseteq \Phi_i} I(Y;S_i) \\
&\geq \frac{1}{N}\sum_{i\in[N_K]}\max_{S_{i,1}\subseteq \Phi_g} I(Y;S_{i,1}) + \frac{K}{N}\delta + \frac{1}{N}\sum_{i\notin[N_K]}\max_{S_i\subseteq \Phi_g} I(Y;S_i) \\
&= \frac{1}{N}\sum_{i\in[N_K]}I(Y;\Phi_g^{\star}) + \frac{K}{N}\delta + \frac{1}{N}\sum_{i\notin[N_K]}I(Y;\Phi_g^{\star}) \\
&= \frac{1}{N}\sum_{i=1}^{N} I(Y;\Phi_g^{\star}) + p\delta.
\end{align*}
\textit{Proof} ends.

As discussed in the main content, the Theorem 1 demonstrates the superiority of personalized invariance on improving the average prediction performance in OOD and Non-IID federated setting, compared with the global invariance which is elicited by the distributional implementation of existing invariant learning methods. In paticular, the improvement $p\delta$ is independent of the number of clients and $p$ is the percentage of the clients with heterogeneous invariance. In other words, $p$ can represent the extent of data heterogeneity across the clients. Therefore, we can conclude that the performance improvement caused by the proposed personalized invariance can be enlarged as the herteroheneity of invariance among the clients increases.

\subsection{Proof of Theorem 2}
\textbf{Theorem 2} Suppose the IRM loss function $\mathcal{L}_{IRM}(\theta;e_j^i)$ is $L_F$-smooth, $\forall$ $i$, $j$, and Assumption 2 holds. If $\gamma \leq \frac{\gamma_0}{\alpha R}$ $\forall$ $\alpha\leq 1$, $R\geq 1$ and $\gamma_0:=\frac{\alpha}{\sqrt{32(R+3)}L_F}$. Then we have
\begin{enumerate}
\item The convergence rate of the global model is given by: \\
	  \begin{small}
	  \begin{equation*}
	  \mathbb{E}\big[{\lVert \nabla\mathcal{L}(\nu_{t^{\star}}) \rVert}^2 \big] \leq \mathcal{O}\big( \mathbb{E}\big[{\lVert \nabla\mathcal{L}(\nu_{t^{\star}}) \rVert}^2\big] \big) := \mathcal{O}\Big( \frac{\Delta_\mathcal{L}R^{\frac{1}{2}}L_F}{\beta T} + \frac{{(\Delta_\mathcal{L}L_F)}^{\frac{3}{4}} {(R\delta_L^2)}^{\frac{1}{4}} }{\alpha^{\frac{1}{2}} T^{\frac{3}{4}}} + \frac{{(\Delta_\mathcal{L} \delta_L L_F)}^{\frac{2}{3}}}{\alpha^{\frac{2}{3}} T^{\frac{2}{3}}} \Big).
	  \end{equation*}
	  \end{small}
\item And, the convergence rate of the personalized models is given by: \\
	  \begin{small}
	  \begin{equation*}
	  \frac{1}{N}\sum_{i=1}^N\mathbb{E}\big[{\lVert \theta_i^{S}(\nu_{t^{\star}}) - \nu_{t^{\star}} \rVert}^2\big] \leq \mathcal{O}\Big( \mathbb{E}\big[{\lVert \nabla\mathcal{L}(\nu_{t^{\star}}) \rVert}^2\big] \Big) + C,
	  \end{equation*}
	  \end{small}
\end{enumerate} where $\Delta_\mathcal{L}:=\mathcal{L}(\nu_0) - \mathcal{L}(\nu_T)$, $C$ is a finite positive constant, and $t^{\star}$ is uniformly sampled from $\{0, 1, ..., T-1\}$. The constant $C$ exists since the number of local iterations $S$ is finite. 

\textit{Proof:} We know the local update of global model follows
\begin{equation}
\nu_{i,r+1}^t = \nu_{i,r}^t - \gamma\underbrace{\nabla\mathcal{L}_{IRM}(\nu_{i,r}^t)}_{=:g_{i,r}^t}
\end{equation}
and the global update is
\begin{align*}
\nu_{t+1} &= \nu_t - \alpha(\nu_t - \frac{1}{N}\sum_{i=1}^N\nu_{i,R}^t) \\
&= \nu_t - \frac{\alpha}{N}\sum_{i=1}^N(\nu_t - \nu_{i,R}^t) \\
&= \nu_t - \underbrace{\alpha\gamma R}_{=:\hat{\gamma}}\underbrace{\frac{1}{NR}\sum_{i=1}^N\sum_{r=0}^{R-1}g_{i,r}^t}_{=:g_t},
\end{align*}
Since the loss function is $L$-smooth, we have
\begin{align*}
&\mathbb{E}[\mathcal{L}(\nu_{t+1}) - \mathcal{L}(\nu_t)] \\
&\leq \mathbb{E}[\left\langle \nabla\mathcal{L}(\nu_t), \nu_{t+1}-\nu_t \right\rangle] + \frac{L_F}{2}\mathbb{E}[{\lVert \nu_{t+1}-\nu_t \rVert}^2] \\
&= -\hat{\gamma}\mathbb{E}[\left\langle \nabla\mathcal{L}(\nu_t), g_t \right\rangle] + \frac{\hat{\gamma}^2 L_F}{2}\mathbb{E}[{\lVert g_t \rVert}^2] \\
&= -\hat{\gamma}\mathbb{E}[{\lVert \nabla\mathcal{L}(\nu_t) \rVert}^2] - \hat{\gamma}\mathbb{E}[\left\langle \nabla\mathcal{L}(\nu_t), g_t-\nabla\mathcal{L}(\nu_t) \right\rangle] + \frac{\hat{\gamma}^2 L_F}{2}\mathbb{E}[{\lVert g_t \rVert}^2] \\
&\leq -\hat{\gamma}\mathbb{E}[{\lVert \nabla\mathcal{L}(\nu_t) \rVert}^2] + \frac{\hat{\gamma}}{2}\mathbb{E}[{\lVert \nabla\mathcal{L}(\nu_t) \rVert}^2] + \frac{\hat{\gamma}}{2}\mathbb{E}[\lVert \frac{1}{NR}\sum_{i,r}^{N,R}g_{i,r}^t-\nabla\mathcal{L}(\nu_t) \rVert^2] + \frac{\hat{\gamma}^2 L_F}{2}\mathbb{E}[{\lVert g_t \rVert}^2] \\
&= -\frac{\hat{\gamma}}{2}\mathbb{E}[{\lVert \nabla\mathcal{L}(\nu_t) \rVert}^2] + \frac{\hat{\gamma}}{2}\mathbb{E}[\lVert \frac{1}{NR}\sum_{i,r}^{N,R}g_{i,r}^t-\nabla\mathcal{L}^i(\nu_t) \rVert^2] + \frac{\hat{\gamma}^2 L_F}{2}\mathbb{E}[{\lVert g_t \rVert}^2] \\
&= -\frac{\hat{\gamma}}{2}\mathbb{E}[{\lVert \nabla\mathcal{L}(\nu_t) \rVert}^2] + \frac{\hat{\gamma}}{2}\mathbb{E}[\lVert \frac{1}{NR}\sum_{i,r}^{N,R}g_{i,r}^t-\nabla\mathcal{L}^i(\nu_t) \rVert^2] \\
&\quad+ \frac{\hat{\gamma}^2 L_F}{2}\mathbb{E}[\lVert \frac{1}{NR}\sum_{i,r}^{N,R}g_{i,r}^t - \nabla\mathcal{L}^i(\nu_t) + \frac{1}{N}\sum_{i=1}^N\nabla\mathcal{L}^i(\nu_t) \rVert^2] \\
&\leq -\frac{\hat{\gamma}}{2}\mathbb{E}[{\lVert \nabla\mathcal{L}(\nu_t) \rVert}^2] + \frac{\hat{\gamma}}{2}\mathbb{E}[\lVert \frac{1}{NR}\sum_{i,r}^{N,R}g_{i,r}^t-\nabla\mathcal{L}^i(\nu_t) \rVert^2] \\
&\quad+ \frac{\hat{\gamma}^2 L_F(1+R)}{2}\mathbb{E}[\lVert \frac{1}{NR}\sum_{i,r}^{N,R}g_{i,r}^t-\nabla\mathcal{L}^i(\nu_t) \rVert^2] + \frac{\hat{\gamma}^2 L_F(1+R)}{2R}\mathbb{E}[{\lVert \nabla\mathcal{L}(\nu_t) \rVert}^2] \\
&\leq -\frac{\hat{\gamma}}{2}(1-\frac{(R+1)\hat{\gamma}L_F}{R})\mathbb{E}[{\lVert \nabla\mathcal{L}(\nu_t) \rVert}^2] + \frac{\hat{\gamma}[1+(1+R)\hat{\gamma}L_F]}{2}\frac{1}{NR}\sum_{i,r}^{N,R}\mathbb{E}[\lVert g_{i,r}^t - \nabla\mathcal{L}^i(\nu_t) \rVert^2]
\end{align*}
Because 
\begin{equation}
\mathbb{E}[\lVert g_{i,r}^t - \nabla\mathcal{L}^i(\nu_t) \rVert^2] = \mathbb{E}[\lVert \nabla\mathcal{L}^i(\nu_{i,r}^t) - \nabla\mathcal{L}^i(\nu_t) \rVert^2] \leq L_F^2\mathbb{E}[\lVert \nu_{i,r}^t - \nu_t \rVert^2],
\end{equation} we first deal with $\mathbb{E}[\lVert \nu_{i,r}^t - \nu_t \rVert^2]$.
We can figure out that
\begin{align*}
&\mathbb{E}[\lVert \nu_{i,r}^t - \nu_t \rVert^2] \\
&= \mathbb{E}[\lVert \nu_{i,r-1}^t - \nu_t - \gamma\nabla\mathcal{L}^i(\nu_t) + \gamma\nabla\mathcal{L}^i(\nu_t) - \gamma g_{i,r-1}^t \rVert^2] \\
&\leq (1+\frac{1}{R})\mathbb{E}[\lVert \nu_{i,r-1}^t - \nu_t - \gamma\nabla\mathcal{L}^i(\nu_t)\rVert^2] + (1+R)\gamma^2\mathbb{E}[\lVert g_{i,r-1}^t - \nabla\mathcal{L}^i(\nu_t)\rVert^2] \\
&\leq (1+\frac{1}{R})(1+\frac{1}{2R})\mathbb{E}[\lVert \nu_{i,r-1}^t - \nu_t \rVert^2] + (1+\frac{1}{R})(1+2R)\gamma^2\mathbb{E}[\lVert \nabla\mathcal{L}^i(\nu_t)\rVert^2] \\
&\quad+ (1+R)\gamma^2 L_F^2 \mathbb{E}[\lVert \nu_{i,r-1}^t - \nu_t \rVert^2] \\
&= (1+\frac{1}{R})(1+\frac{1}{2R}+R\gamma^2 L_F^2)\mathbb{E}[\lVert \nu_{i,r-1}^t - \nu_t \rVert^2] + (1+\frac{1}{R})(1+2R)\gamma^2\mathbb{E}[\lVert \nabla\mathcal{L}^i(\nu_t)\rVert^2]
\end{align*}
Suppose $\gamma^2<\frac{1}{2R^2 L_F^2}$, then we can get
\begin{align*}
&\mathbb{E}[\lVert g_{i,r}^t - \nabla\mathcal{L}^i(\nu_t) \rVert^2] \\
&\leq L_F^2\mathbb{E}[\lVert \nu_{i,r}^t - \nu_t \rVert^2] \\
&\leq R(R+1)\{ [(1+\frac{1}{R})^2]^r - 1\}\gamma^2 L_F^2 \mathbb{E}[\lVert \nabla\mathcal{L}^i(\nu_t)\rVert^2] \\
&\leq R(R+1)[(1+\frac{1}{R})^2]^r \gamma^2 L_F^2 \mathbb{E}[\lVert \nabla\mathcal{L}^i(\nu_t)\rVert^2] \\
\end{align*}
Thus, with Assumption 2, we can get
\begin{align*}
&\frac{1}{NR}\sum_{i,r}^{N,R}\mathbb{E}[\lVert g_{i,r}^t - \nabla\mathcal{L}^i(\nu_t) \rVert^2] \\
&\leq \frac{1}{NR}\sum_{i,r}^{N,R}R(R+1)\gamma^2 L_F^2\mathbb{E}[\lVert\nabla\mathcal{L}^i(\nu_t) \rVert^2] [(1+\frac{1}{R})^2]^r \\
&\leq \frac{1}{N}\sum_{i=1}^N(R+1)\gamma^2 L_F^2 \frac{(e^2-1)}{2R+1}\mathbb{E}[\lVert\nabla\mathcal{L}^i(\nu_t) \rVert^2] \\
&\leq 8R^2\gamma^2 L_F^2 \frac{1}{N}\sum_{i=1}^N \mathbb{E}[\lVert\nabla\mathcal{L}^i(\nu_t) \rVert^2] \\
&= 8R^2\gamma^2 L_F^2 \frac{1}{N}\sum_{i=1}^N \mathbb{E}[\lVert\nabla\mathcal{L}^i(\nu_t) - \nabla\mathcal{L}(\nu_t) + \nabla\mathcal{L}(\nu_t) \rVert^2] \\
&\leq 8R^2\gamma^2 L_F^2 \frac{1}{N}\sum_{i=1}^N \big\{ 2\mathbb{E}[\lVert\nabla\mathcal{L}^i(\nu_t) - \nabla\mathcal{L}(\nu_t) \rVert^2] + 2\mathbb{E}[\lVert \nabla\mathcal{L}(\nu_t) \rVert^2] \big\} \\
&\leq 16R^2\gamma^2 L_F^2 \frac{1}{N}\sum_{i=1}^N\mathbb{E}[\lVert\nabla\mathcal{L}^i(\nu_t) - \nabla\mathcal{L}(\nu_t) \rVert^2] + 16R^2\gamma^2 L_F^2 \mathbb{E}[\lVert \nabla\mathcal{L}(\nu_t) \rVert^2] \\
&\leq 16R^2\gamma^2 L_F^2 \delta_L^2 + 16R^2\gamma^2 L_F^2 \mathbb{E}[\lVert \nabla\mathcal{L}(\nu_t) \rVert^2] \\
\end{align*}
Therefore, we can write
\begin{align*}
&\mathbb{E}[\mathcal{L}(\nu_{t+1}) - \mathcal{L}(\nu_t)] \\
&\leq -\frac{\hat{\gamma}}{2}\big( 1-\frac{(1+R)\hat{\gamma}L_F}{R} \big)\mathbb{E}[\lVert \nabla\mathcal{L}(\nu_t) \rVert^2] \\
&\quad+ \frac{\hat{\gamma}[1+(1+R)\hat{\gamma}L_F]}{2}\big\{ 16R^2\gamma^2 L_F^2 \delta_L^2 + 16R^2\gamma^2 L_F^2\mathbb{E}[\lVert \nabla\mathcal{L}(\nu_t) \rVert^2] \big\} \\
&= -\frac{\hat{\gamma}}{2}\Big\{ 1-\frac{(R+1)\hat{\gamma}L_F}{R}-\frac{16[1+(1+R)\hat{\gamma}L_F]\hat{\gamma}^2 L_F^2}{\alpha^2} \Big\}\mathbb{E}[\lVert \nabla\mathcal{L}(\nu_t) \rVert^2] \\
&\quad+ \frac{8\hat{\gamma}[1+(1+R)\hat{\gamma}L_F]\hat{\gamma}^2 L_F^2 \delta_L^2}{\alpha^2} \\
\end{align*}
Let $\gamma \leq \frac{\gamma_0}{\alpha R}$ $\forall$ $\alpha\leq 1$, $R\geq 1$ and $\gamma_0:=\frac{\alpha}{\sqrt{32(R+3)}L_F}$, then $1-\frac{(R+1)\hat{\gamma}L_F}{R}-\frac{16[1+(1+R)\hat{\gamma}L_F]\hat{\gamma}^2 L_F^2}{\alpha^2} > \frac{1}{2}$. Furthermore, we can get
\begin{align*}
&\mathbb{E}[\mathcal{L}(\nu_{t+1}) - \mathcal{L}(\nu_t)] \\
&\leq -\frac{\hat{\gamma}}{4}\mathbb{E}[\lVert \nabla\mathcal{L}(\nu_t) \rVert^2] + \frac{8\hat{\gamma}^3 L_F^2 \delta_L^2}{\alpha^2} + \frac{8(1+R)\hat{\gamma}^4 L_F^3 \delta_L^2}{\alpha^2}
\end{align*}
That is
\begin{equation*}
\mathbb{E}[\lVert \nabla\mathcal{L}(\nu_t) \rVert^2] \leq \frac{4\big\{ \mathbb{E}[\mathcal{L}(\nu_{t}) -\mathcal{L}(\nu_{t+1})] \big\}}{\hat{\gamma}} + \frac{32\hat{\gamma}^2 L_F^2 \delta_L^2}{\alpha^2} + \frac{32(1+R)\hat{\gamma}^3 L_F^3 \delta_L^2}{\alpha^2}.
\end{equation*}
We can get
\begin{align*}
&\frac{1}{T}\sum_{t=0}^{T-1}\mathbb{E}[\lVert \nabla\mathcal{L}(\nu_t) \rVert^2] \leq \frac{4\sum_{t=0}^{T-1}\mathbb{E}[\mathcal{L}(\nu_{t}) - \mathcal{L}(\nu_{t+1})]}{\hat{\gamma}} + \frac{32\hat{\gamma}^2 L_F^2 \delta_L^2}{\alpha^2} + \frac{32(1+R)\hat{\gamma}^3 L_F^3 \delta_L^2}{\alpha^2}. \\
\quad &= \frac{C_1}{\hat{\gamma}T} + \frac{C_2\hat{\gamma}^2}{\alpha^2} + \frac{C_3\hat{\gamma}^3}{\alpha^2},
\end{align*} where $C_1=4(\mathcal{L}(\nu_0) - \mathcal{L}(\nu_T))$, $C_2=32\delta_L^2 L_F^2$ and $C_3=32(1+R)\delta_L^2 L_F^3$.

We consider the following two cases:
\begin{itemize}
	\item When $\gamma_0 \geq \min\Big\{ \Big(\frac{C_1\alpha^2}{C_3T}\Big)^{\frac{1}{4}}, \Big(\frac{C_1\alpha^2}{C_2T}\Big)^{\frac{1}{3}} \Big\}$, we choose $\hat{\gamma} = \min\Big\{ \Big(\frac{C_1\alpha^2}{C_3T}\Big)^{\frac{1}{4}}, \Big(\frac{C_1\alpha^2}{C_2T}\Big)^{\frac{1}{3}} \Big\}$, we have
	\begin{equation*}
	\frac{1}{2T}\sum_{t=0}^{T-1}\mathbb{E}[\lVert \nabla\mathcal{L}(\nu_t) \rVert^2] \leq \frac{C_1^{\frac{3}{4}}C_3^{\frac{1}{4}}}{\alpha^{\frac{1}{2}}T^{\frac{3}{4}}} + \frac{C_1^{\frac{2}{3}}C_2^{\frac{1}{3}}}{\alpha^{\frac{2}{3}}T^{\frac{2}{3}}}.
	\end{equation*}
	\item When $\gamma_0 \leq \min\Big\{ \Big(\frac{C_1\alpha^2}{C_3T}\Big)^{\frac{1}{4}}, \Big(\frac{C_1\alpha^2}{C_2T}\Big)^{\frac{1}{3}} \Big\}$, we can choose $\hat{\gamma}=\gamma_0$. We can get
	\begin{equation*}
	\frac{1}{2T}\sum_{t=0}^{T-1}\mathbb{E}[\lVert \nabla\mathcal{L}(\nu_t) \rVert^2] \leq \frac{C_1}{\gamma_0 T} + \frac{C_1^{\frac{3}{4}}C_3^{\frac{1}{4}}}{\alpha^{\frac{1}{2}}T^{\frac{3}{4}}} + \frac{C_1^{\frac{2}{3}}C_2^{\frac{1}{3}}}{\alpha^{\frac{2}{3}}T^{\frac{2}{3}}}.
	\end{equation*}
\end{itemize}
Therefore, we have 
\begin{equation}
\frac{1}{T}\sum_{t=0}^{T-1}\mathbb{E}[\lVert \nabla\mathcal{L}(\nu_t) \rVert^2] \leq \mathcal{O}\Big( \frac{C_1}{\gamma_0 T} + \frac{C_1^{\frac{3}{4}}C_3^{\frac{1}{4}}}{\alpha^{\frac{1}{2}}T^{\frac{3}{4}}} + \frac{C_1^{\frac{2}{3}}C_2^{\frac{1}{3}}}{\alpha^{\frac{2}{3}}T^{\frac{2}{3}}} \Big).
\end{equation}

As regard to the convergence rate of the personalized models, the personalized models are locally updated via
\begin{equation}
\theta_i^{\star}(\nu)\in\argmin_{\theta}\mathcal{L}_{IRM}(\theta;\mathcal{E}_{train}^i) + \beta{\lVert \theta - \nu \rVert}^2.
\end{equation}
When $\mathcal{L}_{IRM}(\theta;\mathcal{E}_{train}^i)$ is $L_F$-smooth, after $S$ local iterations, we have~\citep{dinh20pFedMe}
\begin{equation*}
\mathbb{E}[{\lVert \theta_i^S(\nu) - \theta_i^{\star}(\nu) \rVert}^2] \leq \delta^2,
\end{equation*} where $\delta$ is a finite constant.
We can easily get from the proof of Theorem 2 in~\citep{dinh20pFedMe} that
\begin{equation}
\frac{1}{N}\sum_{i=1}^N\mathbb{E}\big[{\lVert \theta_i^{S}(\nu_{t^{\star}}) - \nu_{t^{\star}} \rVert}^2\big] \leq \mathcal{O}\Big( \mathbb{E}\big[{\lVert \nabla\mathcal{L}(\nu_{t^{\star}}) \rVert}^2\big] \Big) + C,
\end{equation} where $C$ is a positive constant and it exists since the number of local iterations $S$ is finite.

\textit{Proof} ends.

\subsection{Proof of Theorem 3}
Adopting the similar assumptions in~\citep{arjovsky2019IRM}, we first define a structural equation model (SEM) for each client to interpret the generating process of local data. On client $i (i\in[N])$, the data is generated according to:
\begin{align*}
&Y_i^e = H_i^e \cdot \rho_i + \epsilon_i^e, \quad\quad H_i^e \perp \epsilon_i^e, \quad\quad \mathbb{E}[\epsilon_i^e] = 0, \\
&X_i^e = G_i(H_i^e,Z_i^e)
\end{align*}
where $\rho \in \mathbb{R}^c$. The random variables $X_i^e, H_i^e$ and $Z_i^e$ take values in $\mathbb{R}^d, \mathbb{R}^c$ and $\mathbb{R}^l$, respectively. Besides, the $H$ component of $G_i$ is invertible. That is, there exists $\widetilde{G}_i$ satisfying that $\widetilde{G}_i(G_i(h,z)) = h$, for all $h\in\mathbb{R}^c$, $z\in\mathbb{R}^l$. Note that the above assumptions about linearity, statistical independence between the causal variables $H_i^e$ and the noise $\epsilon_i^e$, and zero-mean noise are also required in \textit{IRM}~\citep{arjovsky2019IRM}. 

\textbf{Theorem 3} As discussed in Section~\ref{formulation}, the learning model can be written as $f=\Phi\cdot \omega$. Suppose Assumption~\ref{ass:1} is satisfied. Let the global invariant featurizer $\Phi_g^{\star}\in \mathbb{R}^{d\times d}$ have rank $\tau>0$, and the personalized invariant featurizer $\Phi_i^{\star}$ have rank $\tau + q_i$ where $q_i>0$ $(i\in[N])$ and $d>\max_{i\in[N]} q_i + \tau$. Denote the minimizers of our designed objective by $\{f_{\hat{\theta}_i}=\hat{\Phi}_i \cdot \hat{\omega}_i\}$. Then, if there are at least $d-\tau+\frac{d}{\tau}$ local datasets in $\mathcal{E}^D$ lying in a linear general position of degree $\tau$, 
the following holds on each client $i$ $(i\in[N])$:
\begin{enumerate}
	\item When there are at least $d-(\tau+q_i)+\frac{d}{\tau+q_i}$ training contexts in $\mathcal{E}_{train}^i$ lying in a linear general position of degree $\tau+q_i$, with $\beta$ appropriately chosen, we can guarantee that $\mathbb{E}[Y|\hat{\Phi}_i(X),e] = \mathbb{E}[Y|\hat{\Phi}_i(X),e^{\prime}]$, for all $e, e^{\prime}\in \mathcal{E}_{all}^i$, and the test error
	$\mathbb{E}_{(X,Y)\sim \mathbb{P}^{e}}[\ell(f_{\hat{\theta}_i}(X),Y)] = \min_{\omega, Z\subset \Phi_i}\mathbb{E}_{(X,Y)\sim \mathbb{P}^{e}}[\ell(\omega(Z(X)),Y)]$ for any $e \in \mathcal{E}_{all}^i$. 
	\item Otherwise, with $\beta$ appropriately chosen, we can guarantee that $\mathbb{E}[Y|\hat{\Phi}_i(X),e] = \mathbb{E}[Y|\hat{\Phi}_i(X),e^{\prime}], for\text{ }all\text{ }e, e^{\prime}\in \mathcal{E}_{all}^i$.
\end{enumerate}
\textit{Proof}: 
We denote the minimizer of the objective (\ref{ob:glob-irm}) by
\begin{equation*}
f_{{\nu}^{\star}}\in\argmin_{\nu}\sum_{i=1}^N \mathcal{R}(f_{\nu};D_i) + \lambda\cdot{\lVert \nabla_{\bar{\omega}|\bar{\omega}=1.0}\mathcal{R}(\bar{\omega}\cdot f_{\nu};D_i)\rVert}^2,
\end{equation*} where $f_{\nu^{\star}} := \Phi_{\nu^{\star}}\cdot \omega_{\nu^{\star}}$.

When there are at least $d-\tau+\frac{d}{\tau}$ local datasets in $\mathcal{E}^D$ lying in a linear general position of degree $\tau$, using Theorem 9 in~\citep{arjovsky2019IRM} we can conclude that $\Phi_{\nu^{\star}}$ elicits the global invariant prefictor $f_{\nu^{\star}} = \Phi_{\nu^{\star}}\cdot \omega_{\nu^{\star}}$ for all $e\in\mathcal{E}_{all}$, where $\mathcal{E}_{all}^i\subset \mathcal{E}_{all}$ holds for all $i\in[N]$.

\paragraph{Proof of conclusion 1} \textit{Clue}: we can prove conclusion 1 by constructing a contradiction. \\ 
Firstly, when there are at least $d-(\tau+q_i)+\frac{d}{\tau+q_i}$ training contexts in $\mathcal{E}_{train}^i$ lying in a linear general position of degree $\tau+q_i$, we can conclude that $\hat{\Phi}_i$ elicits the invariant prefictor $f_{\hat{\theta}_i} = \hat{\Phi}_i\cdot \hat{\omega}_i$ for all $e\in\mathcal{E}_{all}^i$. For the test error guarantee,

Suppose there exists a model $f_{\bar{\theta_i}}=\bar{\Phi}_i\cdot\bar{\omega}_i$ that is the minimizer of upper objective (\ref{ob:per-irm}), but there isn't any apppropriate $\beta$ such that $f_{\bar{\theta_i}}$ guarantees the test error $\mathbb{E}_{(X,Y)\sim \mathbb{P}^{e}}[\ell(f_{\bar{\theta}_i}(X),Y)] = \min_{\omega, Z\subset \Phi_i}\mathbb{E}_{(X,Y)\sim \mathbb{P}^{e}}[\ell(\omega(Z(X)),Y)]$ for any $e \in \mathcal{E}_{all}^i$.

As regard to the objective (\ref{ob:per-inv}): $\min_{\theta_i} \sum_{j=1}^{K_i}\mathcal{R}(f_{\theta_i};e_j^i) + \lambda\cdot{\lVert \nabla_{\bar{\omega}|\bar{\omega}=1.0}\mathcal{R}(\bar{\omega}\cdot f_{\theta_i};e_j^i)\rVert}^2 + \beta{\lVert \theta_i - {\nu}^{\star}\rVert}^2$. We know that there always exists an appropriate $\beta$ makes it equivalent to
\begin{align*}
&\min_{\theta_i} \sum_{j=1}^{K_i}\mathcal{R}(f_{\theta_i};e_j^i) + \lambda\cdot{\lVert \nabla_{\bar{\omega}|\bar{\omega}=1.0}\mathcal{R}(\bar{\omega}\cdot f_{\theta_i};e_j^i)\rVert}^2 \\
&s.t. \quad {\lVert \theta_i - {\nu}^{\star}\rVert}^2 \leq \delta_d^2
\end{align*} where $\delta_d$ is a constant. In other word, with appropriate $\beta$ chosen, $f_{\bar{\theta_i}}=\bar{\Phi}_i\cdot\bar{\omega}_i$ can guarantee that $\mathbb{E}_{(X,Y)\sim \mathbb{P}^{e}}[\ell(f_{\bar{\theta}_i}(X),Y)]$ is minimized for all $e\in\mathcal{E}_{all}^i$. This result contradicts the previous assumption.

Therefore, the conclusion 1 in Theorem 2 is proved.

\paragraph{Proof of conclusion 2} When the training contexts on local client is insufficient, the minimizer of the objective (\ref{ob:per-irm}) can necessarily rely on the spurious features. However, with an appropriate $\beta$ chosen, the minimizers can be constrained to be $f_{\hat{\theta}_i}=\hat{\Phi}_i \cdot \hat{\omega}_i = f_{\nu^{\star}}$. It has been proved $f_{\nu^{\star}}$ can guarantee $\mathbb{E}[Y|\Phi_{\nu^{\star}}(X),e] = \mathbb{E}[Y|\Phi_{\nu^{\star}}(X),e^{\prime}], for\text{ }all\text{ }e, e^{\prime}\in \mathcal{E}_{all}$. Because $\mathcal{E}_{all}^i\subset \mathcal{E}_{all}$ holds for all $i\in[N]$, $f_{\nu^{\star}}$ can naturally guarantee that $\mathbb{E}[Y|\Phi_{\nu^{\star}}(X),e] = \mathbb{E}[Y|\Phi_{\nu^{\star}}(X),e^{\prime}], for\text{ }all\text{ }e, e^{\prime}\in \mathcal{E}_{all}^i$. \\

\textit{Proof} ends.

\subsection{Detailed experimental setup}
\label{exp_setup}
In this chapter, we will provide the detailed experimental setup, including the dataset setup, model selection and more implementation details.

\paragraph{Dataset setup} We give the detailed setup of the adopted datasets in Table~\ref{table:data}. For the RC-MNIST dataset, we use the first 50000 images in the train-set of MNIST~\cite{lecun1998MNIST} to constrcut the train-sets for the clients, and use the other 10000 images in the train-set of MNIST~\cite{lecun1998MNIST} to constrcut the test-sets for the clients. Finally, in every trial, each client has 12500 data instances for training and 2500 data instances for testing. In the original MNIST dataset, the shape of the images is $28*28$.  After the construction, the images in the obtained RC-MNIST dataset have the size of $28*28*2$. To save the computation cost, we downsample the images to $14*14*2$ before they are input into the first layer of the neural network. The RC-FMNIST dataset is constructed and processed using the same strategies as the RC-MNIST dataset. The WaterBird dataset is constructed using the bird dataset (Caltech-UCSD Birds-200-2011~\cite{wah2011CUB}) and the background dataset (Places~\cite{zhou2017places}), as first introduced in~\cite{sagawa2019DRO}. We adopt the same law as in~\cite{sagawa2019DRO} to construct our datasets. However, we only use the bird photographs in the train-set of Caltech-UCSD Birds-200-2011 which contains 5994 instances. For each client, we randomly sample 400 instances from the index range (shown in Table~\ref{table:data}) of bird dataset to generate the train-set and 100 instances from the same range to generate the test-set. Besides, all of the corresponding background photographs are randomly sampled from the background dataset (Places~\cite{zhou2017places}) without replacement.
\begin{table}[htb]
  \caption{The detailed setup of the local datasets}
  \label{table:data}
  \centering
  \begin{small}
  \setlength{\tabcolsep}{2.0mm}{
  \begin{tabular}{llcccc}
    \toprule
     & & Client 1 & Client 2 & Client 3 & Client 4 \\
    \midrule
    \multirow{4}{*}{RC-MNIST} & \multirow{2}{*}{Train-set} & $p_{train}^e=0.95$ & $p_{train}^e=0.90$ & $p_{train}^e=0.85$ & $p_{train}^e=0.80$ \\
     & & Rotated by $0^{\circ}$ & Rotated by $90^{\circ}$ & Rotated by $180^{\circ}$ & Rotated by $270^{\circ}$\\
    \cmidrule(r){2-6}
     & \multirow{2}{*}{Test-set} & $p_{test}^e=0.10$ & $p_{test}^e=0.10$ & $p_{test}^e=0.10$ & $p_{test}^e=0.10$ \\
     & & Rotated by $0^{\circ}$ & Rotated by $90^{\circ}$ & Rotated by $180^{\circ}$ & Rotated by $270^{\circ}$\\
    \midrule
    \multirow{4}{*}{RC-FMNIST} & \multirow{2}{*}{Train-set} & $p_{train}^e=0.95$ & $p_{train}^e=0.90$ & $p_{train}^e=0.85$ & $p_{train}^e=0.80$ \\
     & & Rotated by $0^{\circ}$ & Rotated by $90^{\circ}$ & Rotated by $180^{\circ}$ & Rotated by $270^{\circ}$\\
    \cmidrule(r){2-6}
     & \multirow{2}{*}{Test-set} & $p_{test}^e=0.10$ & $p_{test}^e=0.10$ & $p_{test}^e=0.10$ & $p_{test}^e=0.10$ \\
     & & Rotated by $0^{\circ}$ & Rotated by $90^{\circ}$ & Rotated by $180^{\circ}$ & Rotated by $270^{\circ}$\\
    \midrule
    \multirow{6}{*}{WaterBird} & \multirow{3}{*}{Train-set} & $p_{train}^e=0.95$ & $p_{train}^e=0.90$ & $p_{train}^e=0.85$ & $p_{train}^e=0.80$ \\
     & & Birds sampled & Birds sampled & Birds sampled & Birds sampled \\
     & & from $[0, 1715]$ & from $[1716, 2908]$ & from $[2909, 3844]$ & from $[3845, 5993]$ \\
    \cmidrule(r){2-6}
     & \multirow{3}{*}{Test-set} & $p_{test}^e=0.10$ & $p_{test}^e=0.10$ & $p_{test}^e=0.10$ & $p_{test}^e=0.10$ \\
     & & Birds sampled & Birds sampled & Birds sampled & Birds sampled \\
     & & from $[0, 1715]$ & from $[1716, 2908]$ & from $[2909, 3844]$ & from $[3845, 5993]$ \\
    \bottomrule
  \end{tabular}}
  \end{small}
\end{table}

\paragraph{Model selection} For RC-MNIST and RC-FMNIST dataset, we adopt a deep neural network with two hidden layers (the dimension is 390 and 390 respectively) as the feature extractor and an subsequent fully-connected layer (the dimension is 390) as the classifier. Besides, after each haidden layer, there is a ReLU layer in the feature extractor. In the evaluation on WaterBird dataset, we adopt the standard ResNet-18~\cite{he2016resnet} as learning model. Similarly, the part before the last fully-connected layer works as the feature extractor and the last fully-connected layer works as the classifier, and the output dimension of the feature extractor is 512.

\paragraph{Implementation} 
Besides, the experiments are implemented in PyTorch. We simulate a set of clients and a centralized server on one deep learning workstation (Intel(R) Core(TM) i9-12900K CPU @ 3.50GHz with one NVIDIA GeForce RTX 3090 GPU).

\end{document}